\documentclass[runningheads]{llncs}
\usepackage{amssymb}
\usepackage{graphicx}
\usepackage{url}
\usepackage{color}
\usepackage{bm}

\usepackage{amsmath}
\usepackage{enumitem}
\usepackage{multirow}
\usepackage{subfig}
\usepackage{tikz,forest}
\usepackage{booktabs,arydshln}
\usepackage{pdflscape}
\usepackage{array}
\usepackage{makecell}
\usepackage{caption}
\usepackage{comment}
\usepackage[makeroom]{cancel}

\newcommand{\PreserveBackslash}[1]{\let\temp=\\#1\let\\=\temp}
\newcolumntype{C}[1]{>{\PreserveBackslash\centering}p{#1}}
\newcolumntype{R}[1]{>{\PreserveBackslash\raggedleft}p{#1}}
\newcolumntype{L}[1]{>{\PreserveBackslash\raggedright}p{#1}}

\usepackage{tikz}

\definecolor{darkblue}{rgb}{0.0, 0.0, 0.55}
\definecolor{darkcyan}{rgb}{0.0, 0.55, 0.55}

\usepackage[colorinlistoftodos]{todonotes} % B
\usepackage[colorlinks=true, allcolors=blue]{hyperref}

\definecolor{orange}{HTML}{FFC17D}
\definecolor{green}{HTML}{A1D68B}
\definecolor{lightgray}{HTML}{E8E8E8}

\usepackage{hyperref} % redefinition needed to work with todo notes

\begin{document}

% \address[a]{UOC, Universitat Operta de Catalunya, Spain}
% \address[b]{Dept. of Computational Design, School of Architecture, Carnegie Mellon University, Pittsburgh, US}
% \address[c]{Dept. of Computer Science and Artificial Intelligence, 
% Andalusian Research Institute in Data Science and Computational Intelligence (DaSCI), University of Granada, Spain}

%\newpage
%\begin{frontmatter}
%\maketitle              % typeset the header of the contribution
%\titlerunning{Title short}
\title{Capabilities, Limitations and Challenges of Style Transfer with CycleGANs: % for
A Study on Automatic Ring Design Generation}

\author{Tomas Cabezon Pedroso \inst{1}\and Javier Del Ser\inst{2,3} \and Natalia  D\'iaz-Rodr\'iguez\inst{4}}

\authorrunning{T. Cabezon Pedroso et al.}
\titlerunning{Capabilities and Challenges of Style Transfer with CycleGANs}

% First names are abbreviated in the running head.
% If there are more than two authors, 'et al.' is used.
%

\institute{Carnegie Mellon University, 5000 Forbes Ave., Pittsburgh, PA 15213, USA \and 
TECNALIA, Basque Research \& Technology Alliance (BRTA), 48160 Derio, Spain \and 
University of the Basque Country (UPV/EHU), 48013 Bilbao, Spain \and 
Department of Computer Sciences and Artificial Intelligence, Andalusian Research Institute in Data Science and Computational Intelligence (DaSCI), CITIC, University of Granada, Granada, Spain\\
\email{tcabezon@andrew.cmu.edu, javier.delser@tecnalia.com, nataliadiaz@ugr.es}}

\maketitle
%\cnat{Tomas haz visibles los autores e indica en footnote Corresponding author:  si es el que seguiras teniendo acceso, inetnta sea institucional, y de larga duracion. sino el de la UOC}
\begin{abstract} 

Rendering programs have changed the design process completely as they permit to see how the products will look before they are fabricated. However, the rendering process is complicated and takes a significant amount of time, not only in the rendering itself but in the setting of the scene as well. Materials, lights and cameras need to be set in order to get the best quality results. Nevertheless, the optimal output may not be obtained in the first render. This all makes the rendering process a tedious process.
Since Goodfellow et al. introduced Generative Adversarial Networks (GANs) in 2014 \cite{goodfellow2014generative}, they have been used to %obtain computer-generated 
generate computer-assigned synthetic data, from non-existing human faces to medical data analysis or image style transfer. GANs have been used to transfer image textures from one domain to another. However, paired data from both domains was needed. When Zhu et al. introduced the CycleGAN model, %\cite{CycleGAN2017}, 
the elimination of this expensive constraint % changed all. CycleGANs allow 
permitted transforming one image from one domain into another, without the need for paired data.
This work %studies the possibilities 
validates the applicability of CycleGANs on style transfer from an initial sketch to a final render in 2D that represents a 3D design, a step that is paramount in every product design process. We inquiry the possibilities of including CycleGANs as part of the design pipeline, more precisely, applied to the rendering of ring designs. Our contribution entails a crucial part of the process as it allows the customer to see the final product before buying. This work sets a basis for future research, showing the possibilities of GANs in design and establishing a starting point for novel applications to approach crafts design. % discuss implication of the results and approach for crafts design.


\end{abstract}

% \begin{keyword}
% Machine Learning \sep   Deep Learning \sep   Automatic Design \sep 
% Style Transfer \sep   Generative Adversarial Networks \sep  CycleGAN \sep Generative models
% \end{keyword}

%\end{frontmatter}
\setcounter{footnote}{0} 

\section{Introduction}

With the advances on artificial intelligence and deep learning (DL), the capabilities of computation in the field of design and computational creativity have spun. Machines have gone beyond doing what they are programmed to do, and debates spur questioning the creativity of models that, in any case, will not replace, but can definitely save time and assist designers do what they do best, more efficiently, and focusing on what really requires their expertise and skillful effort. % Although not all computer theorists agree on this [2], and actually, the 
Works in the intersection of design and computation are growing and are more relevant than ever. In the last years, engineers, researchers or artists have begun to explore the possibilities of artificial intelligence for creative tasks that can vary from the AI generated music of Arca that sounds in the MOMA's lobby\footnote{\emph{Arca will use AI to soundtrack NYC's Museum of Modern Art}, \url{https://www.engadget.com/2019-10-17-arca-ai-soundtrack-for-nyc-moma.html}.} to the drawings by AARON computer program that can be visited at TATE Museum\footnote{\emph{Untitled Computer Drawing}, by Harold Cohen, 1982, Tate. (n.d.),  \url{https://www.tate.org.uk/art/artworks/cohen-untitled-computer-drawing-t04167}.}.

This work rises in this same intersection of design and technology, design and engineering, design and computation. The aim of this paper is to explore new areas and applications in which computers will change the way we consider design and the role that computers have on it. When this statement is made, often the fear of computers stealing people’s jobs arises, nevertheless, this is not how we conceive this intersection of computers and design. While algorithms will spend time doing repetitive work, designers will be able to focus on what really matters: the users, the emotional feeling needs, innovations, or other needs… The objective is thus to make the most out of it for all agents taking part in the design process, from computer programs to designers. We believe the tools at our disposal cannot determine what we are capable of creating. Instead, AI, as another tool more, should serve to develop new ideas and not to limit the ones that the designer already has.

This work is organized in two parts, a theoretical one and a practical one, both complementary. The first one is motivated by the recent arrival of Generative Adversarial Networks (GAN) that since they were first introduced few years ago, in 2014 [5], have experimented and exponential growth and development. This research will be focus on the study and comprehension of the theory that supports GANs and their components. In the second part of the work, taking into account all of the above, a new tool of image generation is applied to an actual design problem, in this case, the rendering of an example of the XYU ring (finger, jewelry area). This tool will consist on a CycleGAN that taking as an input the sketch of the shape of the ring will generate a 3D object representation or a rendered image of it. 

Although the steps in the design process can differ among authors, the whole process consists of going from a virtual concept or idea to the materialization in a concrete product\footnote{\emph{Design Thinking}, \url{https://hbr.org/2008/06/design-thinking}.}. This process starts with an initial brainstorming and, later, some of the concepts are developed, prototyped and, only after evaluation, the final product is selected. Computers have become fundamental in these last steps allowing designers not only to materialize their ideas with 3D objects and renders but also to show the clients how the final products look like. Actually, the famous furniture seller Ikea reaches their clients with the yearly catalogs, full of not real images but renderings of it\footnote{\emph{Why IKEA Uses 3D Renders vs. Photography for Their Furniture Catalog}, \url{https://www.cadcrowd.com/blog/why-ikea-uses-3d-renders-vs-photography-for-their-furniture-catalog}.}. Our ML model aims to input new tools for this last part of the design process.

While the proposed GANs have been used in a broad set of applications, we limit the scope of this paper to assess a potential technology impact that these models could have in automatic design. Although realistic images could be generated by other means, such as rendering, mocking up or photographing, in this work, the objective is not getting the output images themselves, but rather visually assessing the possibilities and limitations of paired GANs. More particularly, we will use GANs to assess the generative and creative capabilities to construct realistic and physically plausible designs.  The concrete example we take is the rendering of a sketch of the XYU ring example. To approach this issue, we need to understand how GANs create images.

The rest of this paper is organized as follows. Section \ref{sec:relatedwork} briefly describes the state of the art on design tools and generative models for computational creativity with respect to design. Section \ref{sec:proposal} presents the proposed design model pipeline, Section \ref{sec:results} shows and examines critically the obtained results. Finally, Section \ref{sec:discussion} concludes with insights for further research and development.

\section{Related work}
\label{sec:relatedwork}

A large body of works has emerged in the literature exploiting the highly performing abilities of generative models such as GANs. We briefly discuss those more closely related to automatic computational design generation.

%\subsection{Neural rendering}
With respect to neural rendering models, some approaches produce photorealistic renderings given noisy or incomplete 3D or 2D observations. In Thies et al. \cite{thies2018ignor}, incomplete 3D inputs are processed to yield rich scene representations using neural textures, which regularize noisy measurements. Similar to our work, Sitzmann et al. \cite{sitzmann2019deepvoxels} aggregate and encode geometry and appearance into a latent vector that is decoded using a differentiable ray marching algorithm. In contrast with our work, these methods either require 3D information during training, complicated rendering priors or expensive inference schemes. In \cite{dupont2020equivariant} they present a way to learn neural scene representations directly from images, without 3D supervision, which permits to infer and render scenes in real time, while achieving comparable results to models requiring minutes for inference.

Since the introduction of generative adversarial networks (GANs) \cite{goodfellow2014generative} and its spread use to generate data --from images to sound, music or even text--, a \textit{zoo} of GANs has emerged. In the plethora of existing models we focus on style transfer models that generally consist of translating images from one domain to a different one, where some dimension or data generating factor should be preserved. Style transfer was proposed by \cite{gatys2015neural} as a neural algorithm able to disentangle content from style from an artistic image, and recombine these elements being taken from arbitrary images.

Among the most popular models for style transfer there are models that use paired datasets to perform image-to-image translation ~\cite{Isola_2017}. Image-to-image translation models can be used to generate street imaging from semantic segmentation masks (DCGAN~\cite{Radford_2016}, Pix2pixHD~\cite{Wang_2018}, DRPAN~\cite{Wang_2018b}, SPADE~\cite{Park_2019}, or OASIS~\cite{Schonfeld_2021}); however the need for paired data makes data collection tedious and costly.

When high-resolution photorealism is a priority despite the computational cost, models such as Pix2pixHD~\cite{Wang_2018} have demonstrated to generate accurate images that are both physically-consistent and photorealistic (e.g., to visualize the impact of floods or ice melt \cite{lutjens2021physically,lutjens2020physics}). %toDo add upon aceptation lutjens2020physics

Alternative to GANs also exist to learn the distribution of possible image mappings more accurately  ~\cite{Casale_2018}, e.g., normalizing flows~\cite{Rezende_2015,Lugmayr_2020} or variational autoencoders~\cite{Kingma_2013}, although they have shown this happens in detriment of the result realistic effect \cite{Dosovitskiy_2016,Zhu_2017}.

%%%%%%%%%%%%%
\section{Proposed model for automatic ring design generation}
\label{sec:proposal}
In this section we present the motivating design problem and the model proposed to achieve this objective in an automatic manner. 

\subsection{Practical use case: designing XYU rings: traditional ring design pipeline}

XYU is not only a ring but an algorithm to create rings%ToDo add when accepted
\footnote{The XYU ring project is key to understanding this work. More information in \url{https://tomascabezon.com/}. XYU is the name of this project, it is not an acronym, but the name of this ring composed of 3 randomly chosen letters.}\cite{cabezonMtesis}. This algorithm uses splines to generate infinite ring possibilities. The starting point is set by the user, who specifies the number of splines and the length and thickness of the ring. The control points of the splines are randomly selected and adapted to make them continuous on the ring. If the user does not like the result, the algorithm can be run again until an aesthetic shape is achieved. 
\begin{figure}[h!]
    \centering
    \includegraphics[width=0.6\textwidth]{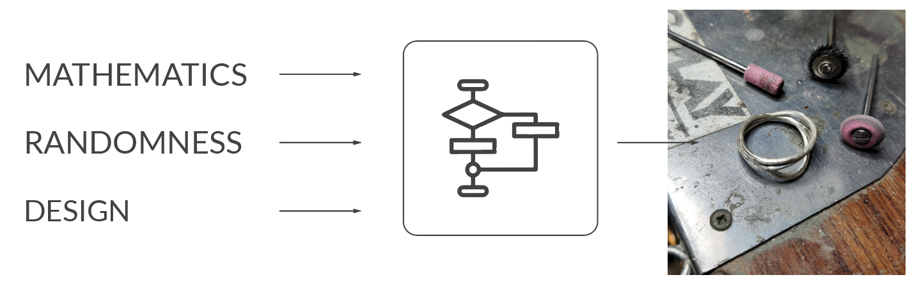} 
    \caption{Original conceptualisation schema for traditional ring design processes.}
    \label{fig:concept schema}
\end{figure}

%\subsection{Traditional Ring Design Pipeline}

The XYU ring algorithm allows the users to design their own 3D ring example based on their preferences, which are used by the algorithm to generate random rings. Each time the XYU ring code is run, a different and unique ring is produced. This procedure permits to personalize each of the XYU ring examples. Once the user finds a ring he/she likes, this is automatically modelled on \textit{Maya} computer graphics application  and the 3D object is sent to the jeweler. Each ring is 3D printed and cast, so each piece is unique. %cast emitido

\begin{figure}[h!]
    \centering
    \includegraphics[width=0.9\textwidth]{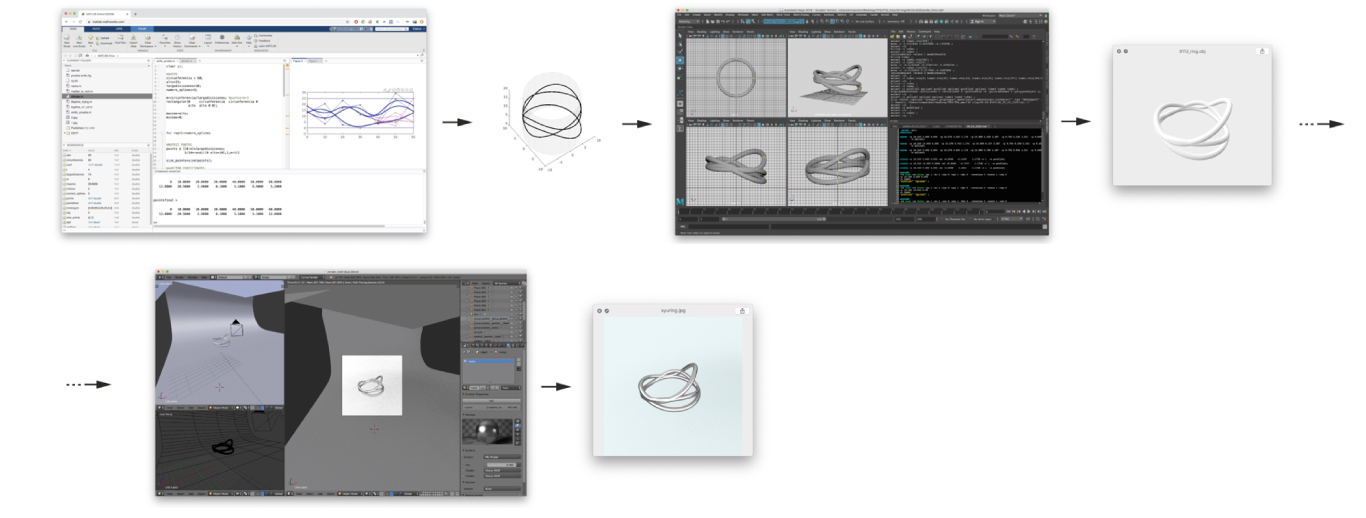} 
    \caption{Traditional pipeline and steps followed. On top Matlab program screenshot where the algorithm is run. In the middle, the Maya application with the 3D object of the ring. On the bottom, Blender software and the rendered image.}
    \label{fig:traditional}
\end{figure}

The description of the actually used programs and different steps followed to go from the initial starting data to the final design of the ring are shown in Figure \ref{fig:traditional}. The algorithm is run on Matlab, the ring is later automatically translated into Maya Embedded Language (MEL) language where the ring is 3D modelled and an .obj/.stl file is created. Finally, the realistic images of the ring are rendered using the Blender software.
%brand = marcas. Extrudir: LENGUAJE TÉCNICO Impeler con una bomba un metal fundido para producir, a través de una matriz adecuada, barras, tubos, varillas y distintas secciones perfiladas.

To generate the 3D object and send the .obj file to the jeweler who will 3D print and cast it, \textit{Maya 3D} application is used. To do so, the information of the ring is passed to Maya using the Maya MEL coding language. This is, the output of the Matlab algorithm is a .txt file with the instructions of the curves that generate the different brands of the ring, the splines, as well as the %circles 
circumferences that will be extruded along the splines to form the 3D object. Therefore, the information to generate the ring in 3D is passed as coded instructions to Maya.

\begin{figure}[h!]
    \centering
    \includegraphics[width=0.7\textwidth]{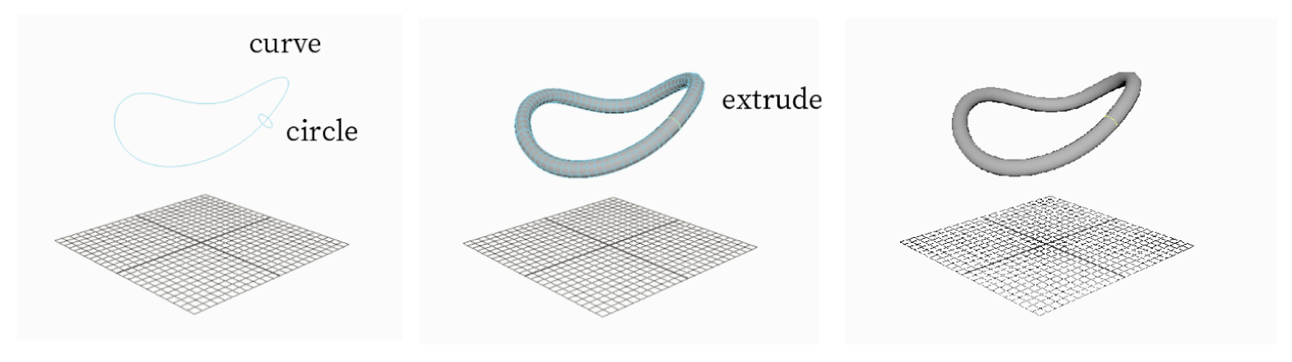} 
    \caption{Schema of how each of the brands of the ring are 3D modelled on Maya 3D modelling software. The points that compose the curve as well as the circle that will be extruded along this curve are passed to the Maya program using Maya MEL coding language.}
    \label{fig:extrusion}
\end{figure}

The last stage of the process corresponds to the rendering of the final product. In this last step, Blender rendering program is used to generate realistic images on the ring and to show the final product to the customer.%\cnat{DONE! explicar la primera vez que mencionas xyu por que se llama xyu?}
\begin{figure}[h!]
    \centering %ToDo make it 0.8 in long version
    \includegraphics[width=0.8\textwidth]{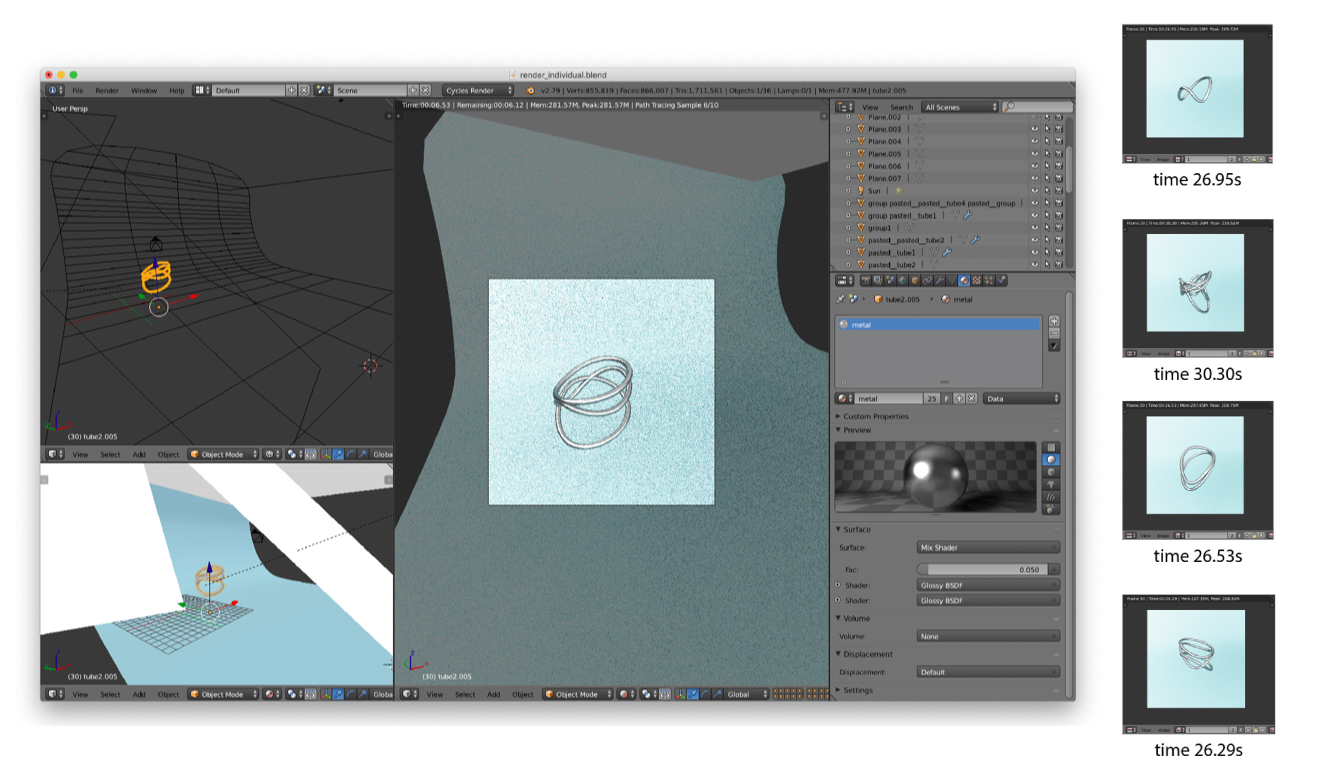} 
    \caption{Rendering settings and image rendering duration times for a set of $1000\times 1000$ pixel size sample images.}
    \label{fig:extrusion}
\end{figure}

\textit{Rendering} is the process of turning a 3D scene into a 2D image. A 3D scene is composed of various elements apart from the object we want to render, such as the background, the camera, the materials and the light. This step is the most tedious part of the XYU ring generation: the rendering of images not only takes a long time to be calculated, but scenes need to be arranged and the images do not always render as expected the first time. As a matter of fact, companies that create whole animation films by rendering each of the video frames of their films, such as Pixar, have rendering directors to optimize this process.

To calculate how long it takes to render an image, on top of the time needed to compute the color of each pixel, which is not the longest one of the process, the scene setting time, the lighting configuration and the material generation and selection times should be added. A properly rendered image of an XYU ring would take, in total, around an hour in the making.

 As it was seen, the original idea of completely automatizing the whole process of the XYU ring generation by the user was not achieved, as intermediate external programs need to be used in the process by the designer. Figure \ref{fig:extrusion} shows the actual process intermediate steps needed for the 3D object generation and rendering.
  \begin{figure}[h!]
    \centering
    \includegraphics[width=0.9\textwidth]{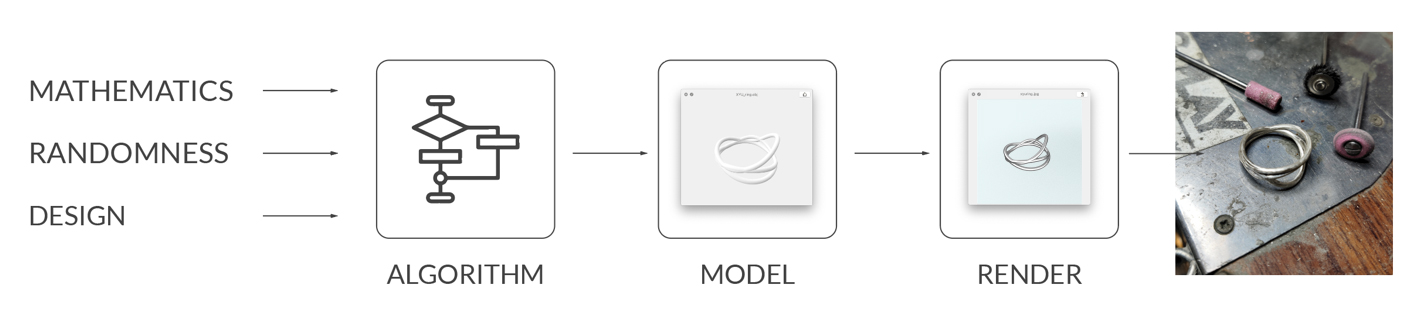} 
    \caption{Current traditional design pipeline schema from left to right: the algorithm run in Matlab, % (coudl be Python too)
    the 3D object modelled in Maya, rendered image in Blender and final product after being made by the jeweler. This process is slow and requires human intervention to render the 3D model.}
    \label{fig:extrusion}
\end{figure}

\subsection{Proposed Pipeline: Automatic design rendering through generative models for image-to-image translation}% with CycleGANs}

The initial process in which the generation of rings has been completely automatized has not been achieved yet. This limitation was the starting point that motivates this work. Therefore, %finding a new approach for this ring design generation algorithm in which no need of the designer has been tackled in this work which has resulted in this proposal of including a CycleGAN in the process that allows the algorithm to generate the rendered images of the ring and show them to the user.
we propose a new approach for this ring design generation algorithm in which there is no need for the designer to be involved in the process of generating rendered images of the ring to show to the user.

 \begin{figure}[h!]
    \centering
    \includegraphics[width=0.6\textwidth]{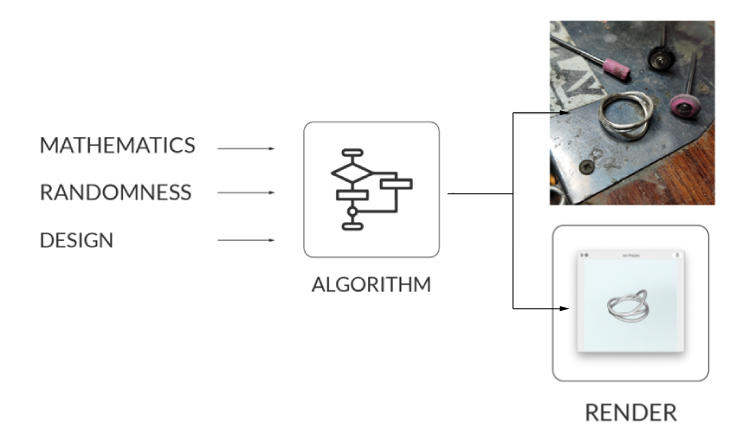}  
    \caption{Proposed model design pipeline. %In this case, 
    The algorithm generates not only the ring 3D object, but also its rendered images. The traditional rendering process is skipped, because it is automatically done by the CycleGAN within the full algorithm. %process (including as in previous Figure, the creation of a 3D object file). 
    The interesting aspect is that this allows: 1) the end user to instantaneously visualize the designed sample, which would not be possible without intervention from the designer to render it. 2) automating the full process. %ToDo important (e.g. using a single programming language such as Python). %, since including the ring generator in the algorithm part, there is no need for additional software programs (e.g. Matlab as in algorithm in Fig. \ref{fig:extrusion})    nor human intervention.
     }
  
    \label{fig:extrusion}
\end{figure}

%Current method in Fig. \ref{fig:extrusion} is slow and also requires human intervention to render the 3D model. Using a CycleGAN this process is completely automated, as well as making the rendering process faster. This is because the designer does not have to be part of the iterative ring generation process.  
It is important to note that the process for the jeweler does not change, as the 3D model file (e.g. .obj, .stl) is in both cases created for it to be 3D printed and later cast by the jeweler. This means the ring sketch is generated by the Matlab algorithm, which generates the different spline curves that compose the ring. These 3D spline curves are plotted in 2D and this image is used as input sketch for the CycleGAN.
%In Fig. \ref{fig:extrusion}, 
Therefore, since the algorithm also produces the rendering of the 3D models, the end user benefits from the CycleGAN by running it, choosing the preferred 3D model, and printing it in 3D, which means the full process is amenable to be automated.

\subsection{CycleGAN as a generative model trained on unpaired images}

CycleGANs \cite{CycleGAN2017} are generative models that are trained on unpaired sets of images in tuple format. They are used to translate between image \textit{styles} or domains. Some examples can be seen in computer vision applications \textit{translating} or transforming images from one domain into another (e.g. \textit{horse2zebra}, \textit{apple2orange}, \textit{photo2Cezanne}, \textit{winter2summer}... and vice versa).

In this work, we propose an innovative use of CycleGANs that consists of speeding up the last parts of the process of the design of XYU rings, i.e., their presentation. To achieve this, a \textit{Sketch2Rendering} CycleGAN is trained. It will get a sketch image as an input and will apply a style transfer to generate a rendered image of the ring. In the following image this CycleGAN can be seen. Our hypothesis is that training a CycleGAN that would render the generated rings would not only ease this process and reduce its time but also automatize and speed up the process.

\begin{figure}[h!]
    \centering
    \includegraphics[width=0.7\textwidth]{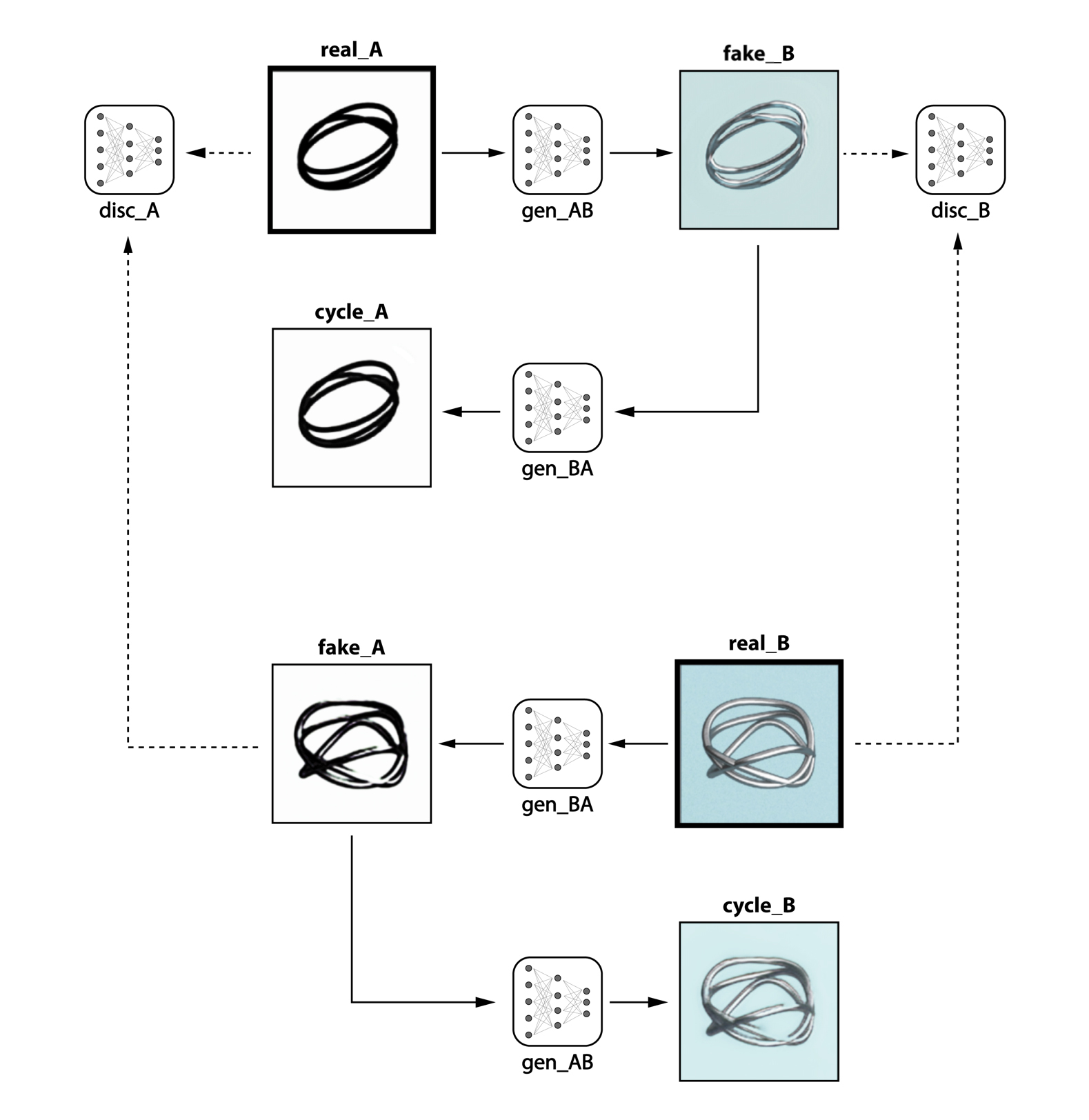} 
    \caption{Proposed CycleGAN where \textit{gen} and \textit{disc} prefixes stand for generator and discriminator, respectively. Input training images are the ones with a thicker %border
    frame.}
    % \ctomas{revisar}\cnat{Duda:Deberian ser la entrada a los diisc las imagenes del mismo anillo? check}
    % \label{fig:proposed cycleGAN}
\end{figure} 

%\subsection{CycleGAN Model}
One reason to choose CycleGANs against other \textit{image2image} translation models is their ability to create infinite samples for a given input. The variational autoencoder (VAE) module that is part of the CycleGAN is %what made us choose 
responsible for this feature, and one of the reasons behind our choice of this architecture. Against other models that are deterministic and thus, limited to produce a single unique output for a given input, a CycleGAN allows to produce a one-to-many output pairings for a given input image.

We propose a CycleGAN based on the one proposed by Zhu et al. %in the paper U
as an unpaired Image-to-Image Translation using Cycle-Consistent Adversarial Networks \cite{CycleGAN2017}. The CycleGAN involves the automatic training of image-to-image translation models without paired examples. This capability is very suitable for this application, as there is no need for paired image datasets, since this is usually challenging and time consuming to obtain.

% MOVED TO APPENDIX.  Indicate if space

\section{Results and analysis}
\label{sec:results}

In this section we present the produced designs by the CycleGAN architecture and provide some visual galleries to analyze them. Models, scripts and notebooks used to produce these results have been made publicly available\footnote{ \url{https://github.com/tcabezon/automatic-ring-design-generation-cycleGAN}, % A notebook to explore and play with the models is in \footnote{
\url{https://tcabezon.github.io/automatic-ring-design-generation-cycleGAN/}}.

\subsection{\textit{Sketch2Rendering} image results}

% \begin{figure}[h!]
%     \centering
%     \includegraphics[width=1.0\textwidth]{figures/image8_.jpg}
%     \caption{Collection of some results of the trained CycleGAN Sketch2Rendering mapping.}
%     \label{fig:Cycle consistency loss}
% \end{figure} 
%\ctomas{he quitado esta imagen}

\begin{figure}[h!]
    \centering
    \includegraphics[width=0.9\textwidth]{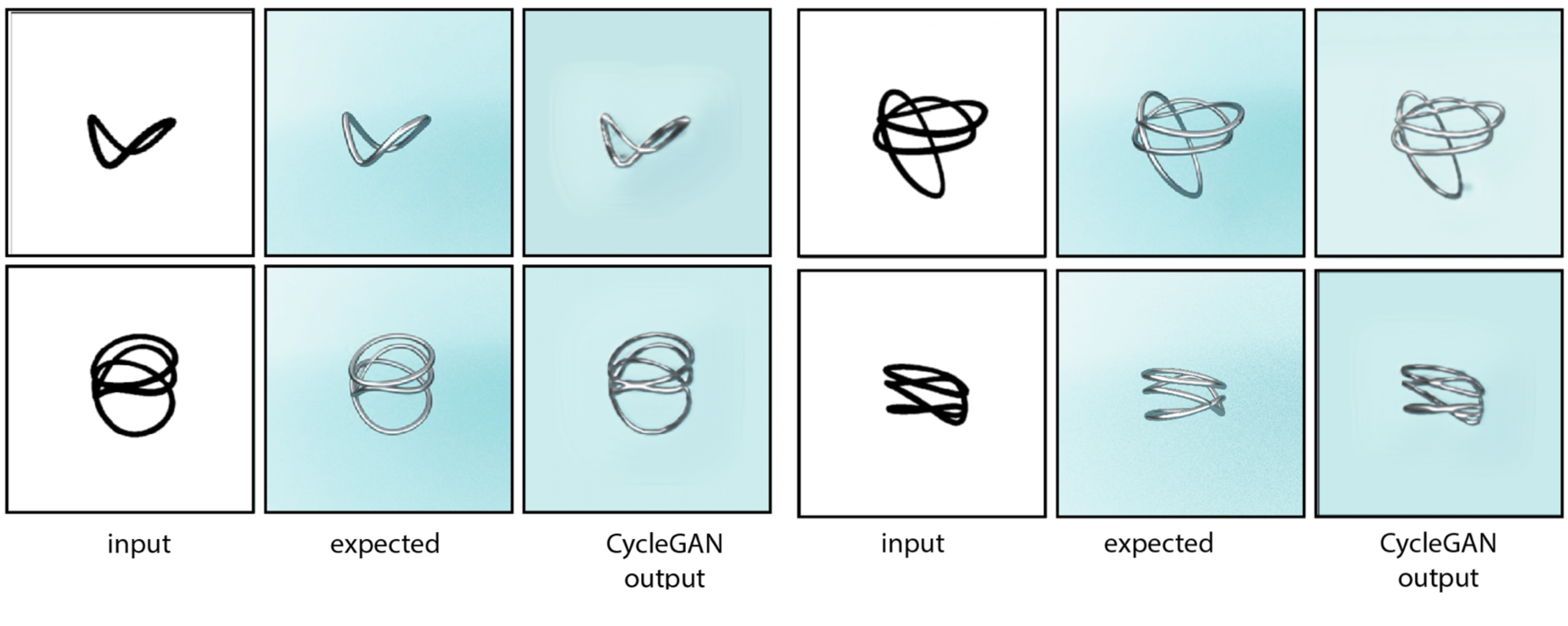} 
    \caption{Different rings in the sketch domain and the rendered domain. The expected rendering were generated with Blender, while the other using the Sketch2Rendering CycleGAN.}
    \label{fig:Cycle consistency loss}
\end{figure}

In Figure \ref{fig:Cycle consistency loss} some examples of the style transfer performed by the once trained \textit{Sketch2Rendering} CycleGAN are shown. For comparison purposes, some of the ring sketches used to train the data have been modelled and rendered using the traditional procedure. In the following image, the outputs of the CycleGAN are show next to what they could be some expected rendered images, using the \textit{Maya} modelling program and \textit{Blender} rendering program. Although the data consists of unpaired images, in Figure  \ref{fig:Cycle consistency loss}, images are show as pairs of the input sketch and the generated image by the Sketch2Rendering CycleGAN and the rendered image using Blender.

\begin{figure}[h!]
    \centering
    \includegraphics[width=0.8\textwidth]{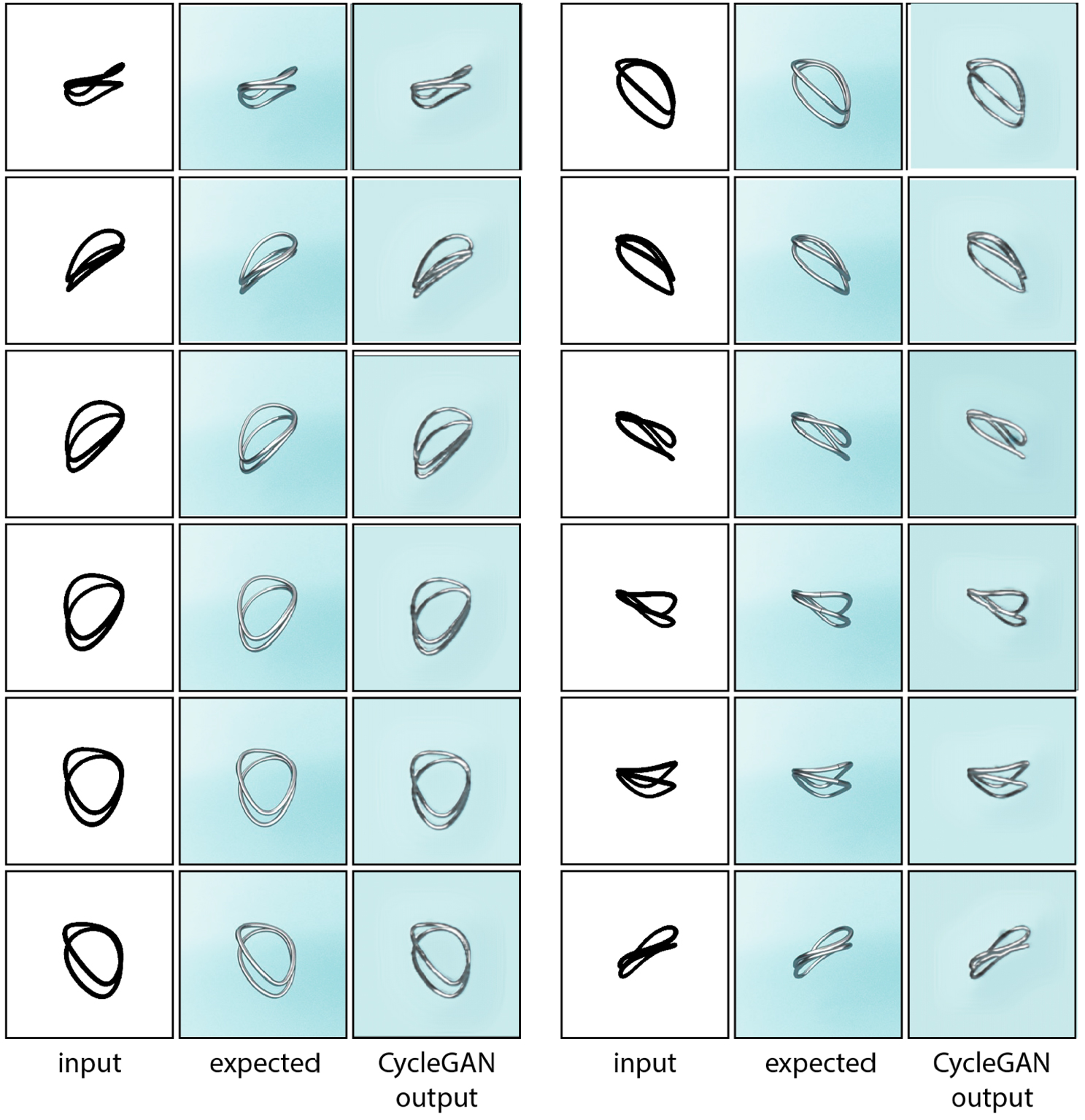} 
    \caption{Different views of the same object in the sketch domain and the rendered domain. The expected rendering was generated with Blender, while the other was generated by the Sketch2Rendering CycleGAN.}
    \label{fig:Cycle consistency loss1}
\end{figure}

\begin{figure}[h!]
    \centering
    \includegraphics[width=0.8\textwidth]{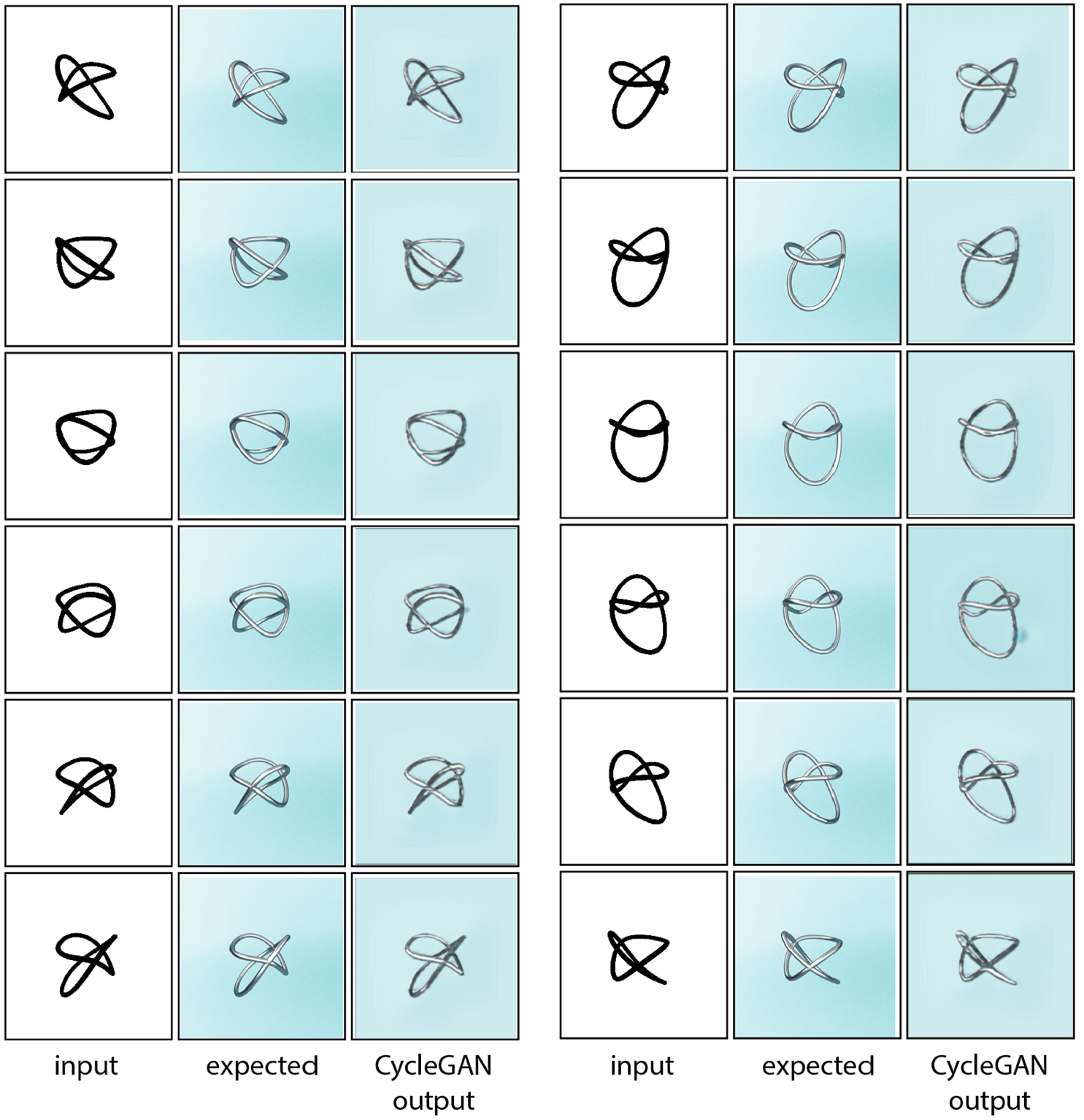} 
    \caption{Different views of the same object in the sketch domain and the rendered domain. The expected rendering was generated with Blender, while the one on the right of it was generated by the Sketch2Rendering CycleGAN. }
    \label{fig:Cycle consistency loss2}
\end{figure}

In Figures \ref{fig:Cycle consistency loss1} and  \ref{fig:Cycle consistency loss2}, 360 degrees of the same XYU ring can be seen. On the left, the input (sketch) image is shown and, next to it, an expected rendered image of the 3D object using Blender. On its right, the output of the \textit{Sketch2Rendering} CycleGAN.

\section{Discussion} \label{sec:discussion}

After having shown the possibilities of the applications of style transfer with CycleGANs for rendering purposes, in this section, the artefacts found during the training and testing, as well as some of the limitations found on the model will be considered.
Although the Sketch2Rendering model can achieve reasonable results in some cases, there are areas for improvement in future works. As it can be seen in the following lines, the results are far from uniformly positive and there are still some challenges and improvements to be done before good quality realistic images of rings are generated by the CycleGAN. The following describes some artefacts found both while training and testing the model.

\subsection{Challenging aspects and detected artefacts}

\subsubsection{Appearance of white spots:}
%\begin
During the training of the model, white blurry spots were found on the output images. These were usually found in the edge of the ring in areas where there is a strong shine on the ring, or where the different bands of the ring intersected.

% \begin{figure}[h!]
%     \centering
%     \includegraphics[width=0.9\textwidth]{figures/image25_.jpg} 
%     \caption{Examples of the white spots artefacts.}
%     \label{fig:Examples of the white spots artefact.}
% \end{figure} 

% \subsubsection{Loss of continuity in some lines in the sketch transformation}

% When the rendered images are transformed into the sketch domain, there is no continuity in the curves that form the ring due to the shiny parts of the rendered image. Although the domain change we are looking for in this work is the sketch-to-rendering change, to train the CycleGAN the whole cycle is applied to the image, so solving this problem with the style transfer from the rendering domain to the sketch domain may be fundamental to obtain better results in the sketch-to-rendering transformation.

% \begin{figure}[h!]
%     \centering
%     \includegraphics[width=0.9\textwidth]{figures/image26_.png}  
%     % \caption{Examples of the of continuity in the rendering-to-sketch domain transformation.}
%      \caption{Examples of lack of continuity in the rendering-to-sketch domain transformation. Some parts of the ring disappear when transformed into the sketch domain.}
% \end{figure} 

%ToDo add in long version \ctomas{creo que si quitamos un artefact, ahorramos bastante espacio}

\subsubsection{\textit{Aureoles} around the foreground object:}% ring}

Exhibiting some kind of background \textit{aureole} around the foreground object may be one of the most commonly found artefacts in the model outputs. %It is a gradient 
The colour gradient or aureole shows around the edges of the ring. We could explain this effect to be due to the different lighting settings, since there is non uniform background color in the full training dataset. Actually, the rendered images created using the Blender program show noise in the background, as if a \textit{Photoshop Film Grain} filter would have been applied to the background. This is due to the renderization parameters on Blender. In order to accelerate the renderization process, the number of calculation steps for the color of each pixel was reduced when the dataset was created. In order to verify whether this is the actual cause of this artefact, in future works, better-quality datasets should be created, not only for the rings themselves but also for the
backgrounds.

\subsubsection{Checkerboard patterns:} Checkerboard effect is a common and one of the most typical artefacts in GANs. The reason for this checkerboard-like pattern in images is due to the upsampling process of the images from the latent space, which becomes visible in images with strong colours. This artefact appears as a consequence of the ability of deconvolutions (i.e., transposed convolutions) to easily show an uneven overlap that adds \textit{more of the metaphorical paint in some places than others} \cite{odena2016deconvolution}. Since there are some solutions proposed for solving such artefact \cite{odena2016deconvolution}, these may be some of the first strategies to be applied in future works to improve the \textit{pixelated}-looking effect on generated images. 

\begin{figure}[h!]
    \centering
    \includegraphics[width=0.7\textwidth]{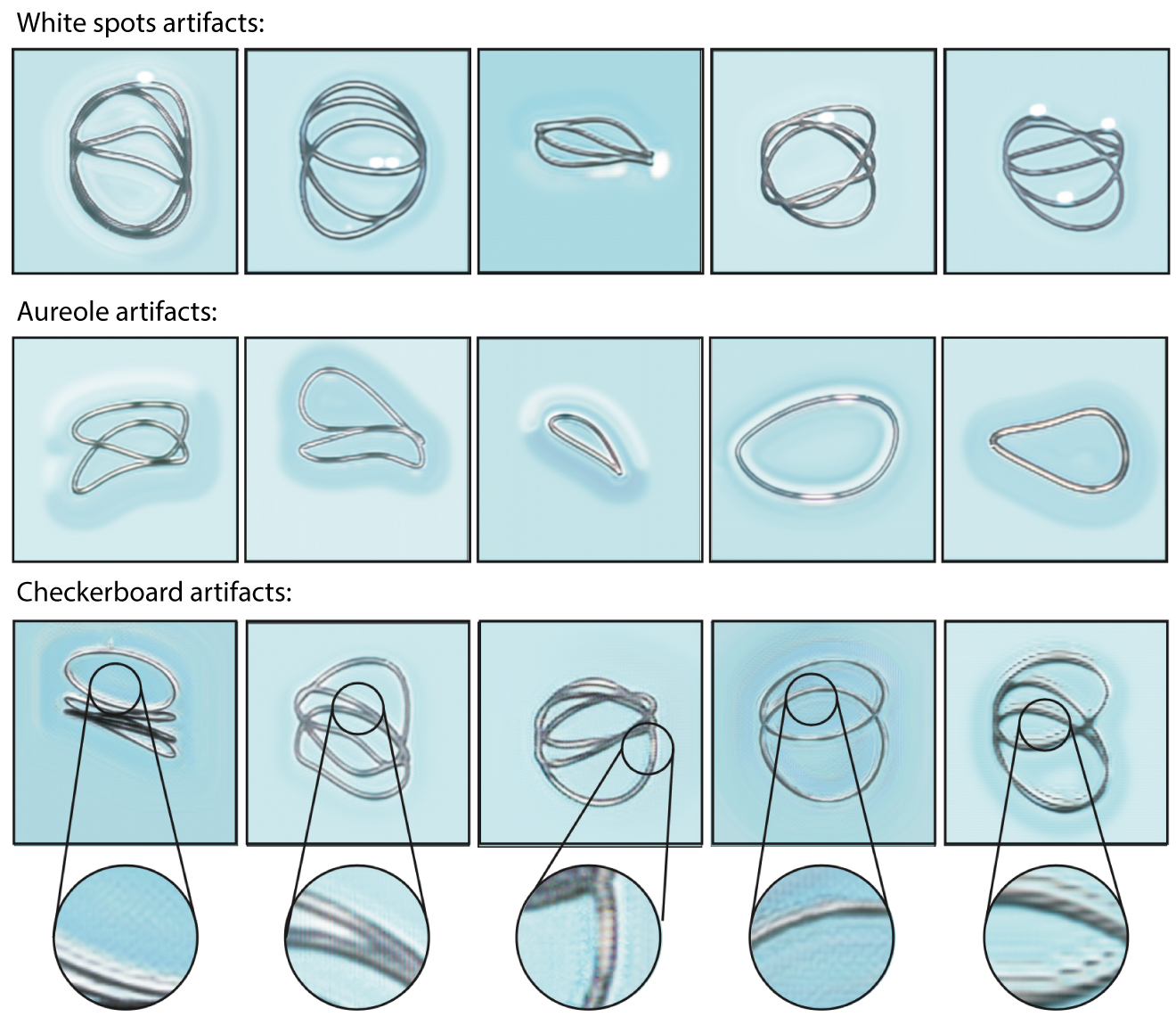}  
    \caption{Examples of generated samples by the Sketch2Rendering model exhibiting the different artefacts. Examples of the white spots artefact on the top, the aureole artefact in the middle, and the checkerboard pattern on the bottom. }
\end{figure} 

%https://github.com/junyanz/pytorch-CycleGAN-and-pix2pix/issues/190
%https://distill.pub/2016/deconv-checkerboard/
%sub-pixels convolutions: https://wandb.ai/wandb/common-ml-errors/reports/How-to-avoid-checkerboard-pattern-in-your-generated-images---VmlldzozNTEzNzk%20https://arxiv.org/abs/2012.00287
% https://www.reddit.com/r/MachineLearning/comments/g9tbge/d_srgan_checkerboard_artifacts/

% \begin{figure}[h!]
%     \centering
%     \includegraphics[width=0.8\textwidth]{figures/image28_.png}  
%     \caption{Examples of generated samples by the Sketch2Rendering model exhibiting the checkerboard artefact.}
    
% \end{figure} 

\subsection{Model limitations}

Apart from the artefacts described in the previous section, some limitations of the actual model have been found. Solving these would %need to change the 
require further %changes in 
development of the model itself, for example, using paired data for the lighting setting %instruction %how would you instruct this? to %know where
to help guide the model to infer where the light comes from, or change how lines in the sketch intersect, %to show which one is on top of the others.
for the model to better disentangle which one is on top of the other.

\subsubsection{Coping with arbitrary lighting settings:}

In the following image, different lighting settings were used in the rendering of the same object with the same materials. Therefore, different images were created. Even if it may not be as realistic, in order to improve the training of the CycleGAN it could be beneficial for the model to be trained %raining if the same rendering settings were always used
using always the same rendering settings, so that the CycleGAN is able to more sharply learn the rendering style. 

Another solution, as previously mentioned, could be using other prior information or labeled data, e.g., adding information about the light position and direction, so the network can learn the differences. However, this would complicate the construction of datasets, since really precise information would be required to make sure that all the information about the lighting settings is included in the labels, which is something to be avoided, due to an increased cost in time and effort of the annotation process. Furthermore, this additional input would require some effective information fusion strategy for the model to leverage this information adequately.

\begin{figure}[h!]
    \centering
    \includegraphics[width=0.8\textwidth]{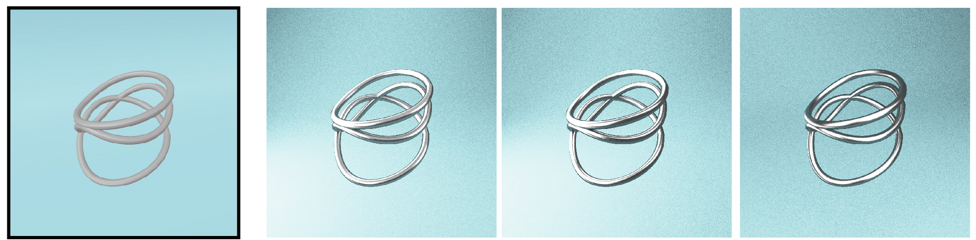}  
    \caption{Examples of the influence of diverse lighting settings on the final renderings generated by the Sketch2Rendering model.}
    \label{fig:lighting.}
\end{figure} 

\subsubsection{Learning to account for a 3D perspective:}

\begin{figure}[h!]
    \centering
    \includegraphics[width=0.9\textwidth]{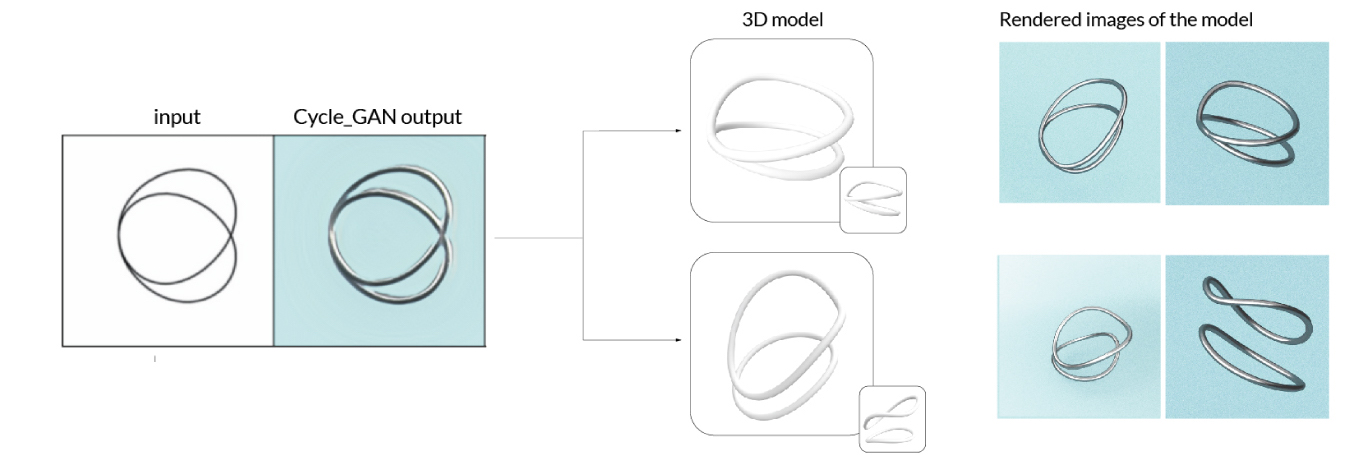}  
    \caption{Example of two lines intersecting in the 2D image, and two different 3D examples of the actually intersecting and non intersecting cases.}
    \label{fig:perspective}
\end{figure}

The sketch input image of the ring is an image of a 3D plot of the different splines that form the ring. Therefore, when in the 2D image two lines intersect, this may be because the lines actually intersect in the 3D space, or it may just be a consequence of the perspective. When creating a plot of a 3D object, some information is lost and thus, there is no way for the CycleGAN to know which of the intersecting lines in the image is on top of which or whether they are actually intersecting. A good way to solve this could be to use a different representation for intersecting lines and those that are not, for example, the diagrams used for knot representation in the study of mathematical knots \cite{murasugi1996knot} could be used. That way the model will learn when to render both of the splines together, or when they are not intersecting. In Figure \ref{fig:perspective} an example of this can be seen, on the top, the splines actually intersect; while in the bottom, both of the splines are separated.

%ToDo: make sure upon aceptation we link the page and repo: el codigo esta aqui: https://github.com/tcabezon/automatic-ring-design-generation-cycleGAN 

% web page en la que poodemos poner el link al archivo mas adelante https://tcabezon.github.io/automatic-ring-design-generation-cycleGAN/ , cuando tenga tiempo pondre mas imagenes de los resultados, y me gustaria poner tambien las datasets. Acabo de darme cuenta de que los modelos entrenados no los pude subir al github porque pesan d

\section{Conclusions and future work}

After presenting the obtained results through the proposed and trained CycleGAN architecture and having discussed the problems encountered, some conclusions around the initially set objectives can be drawn.

First of all, we can assert that the Sketch2Rendering model can achieve compelling results for the rendering of the design through style transfer. That was the initial objective; nevertheless, as it was seen, there are areas for improvement in future works before high quality realistic images of the ring examples are generated by the CycleGAN. However, the results obtained exceed the initial expectations for this work. Indeed this new model supposes a new approach for the XYU ring algorithm and even though perfect results were not obtained, the reduction of the time and the allowance of a complete automation of the different ring design generation, makes this work an exemplary starting point for future research and improvements.

Secondly, this work shows the possibilities lying in the intersection of computation and design, which allows designers to focus on what really matters, while the algorithms do the repetitive non creative work. The rendering style transfer supposes going from the rendering of images that could take up to one hour on the making, to renders generated by the CycleGAN in seconds.

Therefore, it can be concluded that having developed a software that is capable of transferring the rendering style to the initial sketches of the ring, the research objective was achieved. The contribution of this work to the XYU ring design generation algorithm supposes an inflexion point for the way rings are shown to the end user, who now would be able to see real time rendered images of the ring that is being generated while interacting with the algorithm. Although we succeeded at the objective of validating image-to-image translation frameworks for automatic design rendering, some problems were encountered during the development of this work and discussed. We hope the research community finds the potential avenue of future works motivating for the exciting field of computational creativity and AI-assisted design to thrive. 

In the future other types of GANs could be trained, e.g., models for higher resolution such as Pix2Pix, BiGAN or StyleGAN, and train them with paired data when available, to compare the gain in quality with this model's results. This quantitative comparison with other methods could help decide whether more dataset agnostic models that do not require paired data (such as the CycleGANs used in this work), are the best approach for this problem, or instead preparing paired data to train an image-to-image translation GAN is worth the time and effort. %Future works should also assess metrics on some mathematical notions of correctness, and place them in the context of practical implications they could have for jewelry design.
Future works should also assess some mathematical notions of correctness, and practical implications they could have for jewelry design.

%" is unclear, as those issues could just be generic issues that all image-based GAN generations will have."

\section*{Acknowledgments}
Díaz-Rodríguez is supported by %grant
IJC2019-039152-I funded by %MCIN/AEI/10.13039/5011%00011033 
MCIN/AEI/10.13039
/501100011033 by “ESF Investing in your future” and Google Research Scholar Program. 
Del Ser %would like to thank the Basque Government for its funding support through the ELKARTEK program (3KIA project, KK-2020/00049) and the research group MATHMODE (T1294-19). 
is funded by the Basque Government ELKARTEK program (3KIA project, KK-2020/00049) and research group MATHMODE (T1294-19). 

\typeout{}
 \bibliographystyle{unsrt} 
\bibliography{references}

%\newpage
\appendix

\section{APPENDIX: Supplementary materials}

\subsection{Datasets}

 Some randomly selected .jpg images from the different datasets generated for this work are shown in this section. The aim  is to show the diversity of images that have been used for the purpose of training the CycleGAN. 
 
\begin{figure}[h!]
    \centering
    \includegraphics[width=1.0\textwidth]{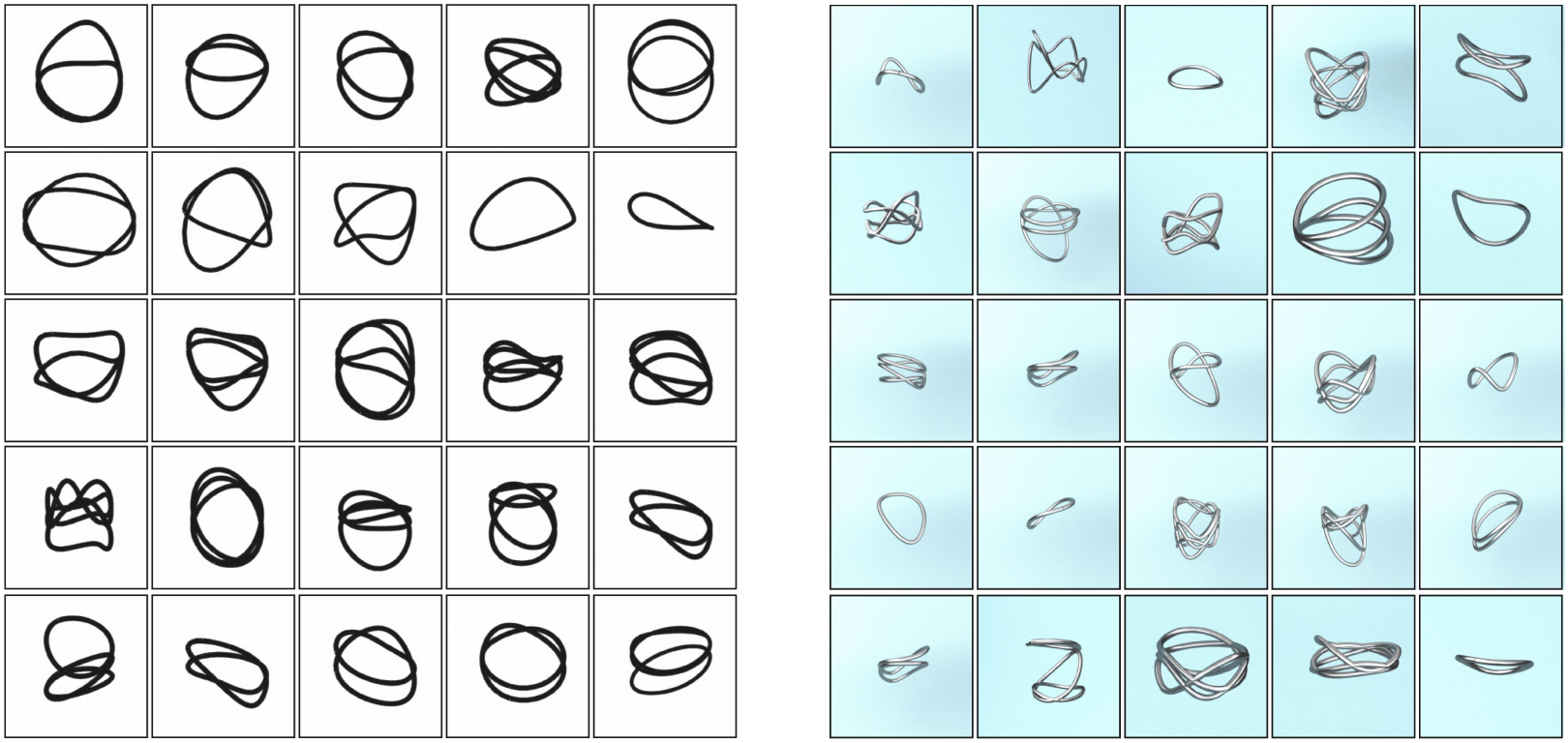}  
    \caption{On the left, random images of wire sketches created in Matlab language and used to train the CycleGAN model (domain A).On the right, random images of the rendered set of images, created using Blender rendering software, and used to train the CycleGAN model (domain B).}
    \label{fig:dataset}
\end{figure}

\paragraph{Sketch2Rendering:} 179 sketch images and 176 rendered images of the training dataset were used for the training. The images were scaled to 400 x 400 pixels when loaded. The Sketch dataset is composed of .jpg images. These, in Figure \ref{fig:dataset}, have been generated using Matlab and the XYU ring algorithm. These images are a 3D plot of the splines that compose each of the rings, all with the same line thickness. The thickness was varied to show different ring thicknesses.

% \begin{figure}[htbp!]
%     \centering
%     \includegraphics[width=0.65\textwidth]{figures/dataset1.png}  
%     \caption{Random images of wire sketches created in Matlab language and used to train the CycleGAN model (domain A).}
%     \label{fig:sketch dataset1}
% \end{figure} 

\paragraph{Rendered dataset:} The images in Figure \ref{fig:dataset} have been created using the Blender rendering software. As it can be seen, although the background color has always been the same blue (\#B9E2EA), the lighting setting has changed, as well as the camera position and orientation, and thus, different shadows and lights can be appreciated across the dataset.

\subsection{CycleGAN Model}

To achieve a cycle consistency among two domains, a CycleGAN requires two generators: the fist generator ($G_{AB}$) will translate from domain A to B, and the second generator ($G_{BA}$) will translate from domain B back to A. Therefore, there will be two losses, one forward cycle consistency loss and other backward cycle consistency loss. These mean that $x*=G_{AB}(G_{BA}(x) )$ and $y*=G_{BA}(G_{AB}(y) )$. % sigue sin estar claro. dar dos pasos de transformacion en ambos. solo cambia el primer dominio de entrada. Por que uno tieine un paso mas? We can think of them as essentially the same, but off by one.\cnat{DONE i meand the are nearly the same but only one step more, X->Y->X->Y... saque la expreesi'on de un libro, lo he comprobado con mi companera de piso y dice que funciona, what you mean?}

\subsubsection{CycleGAN Generator architecture:}

The generator in the CycleGAN has layers that implement three stages of computation:
\begin{enumerate}
    \item The first stage encodes the input via a series of convolutional layers that extract image features.
    \item the second stage then transforms the features by passing them through one or more residual blocks.
    \item The third stage decodes the transformed features using a series of transposed convolutional layers, to build an output image of the same size as the input.
\end{enumerate}

\begin{figure}[h!]
    \centering
    \includegraphics[width=0.8\textwidth]{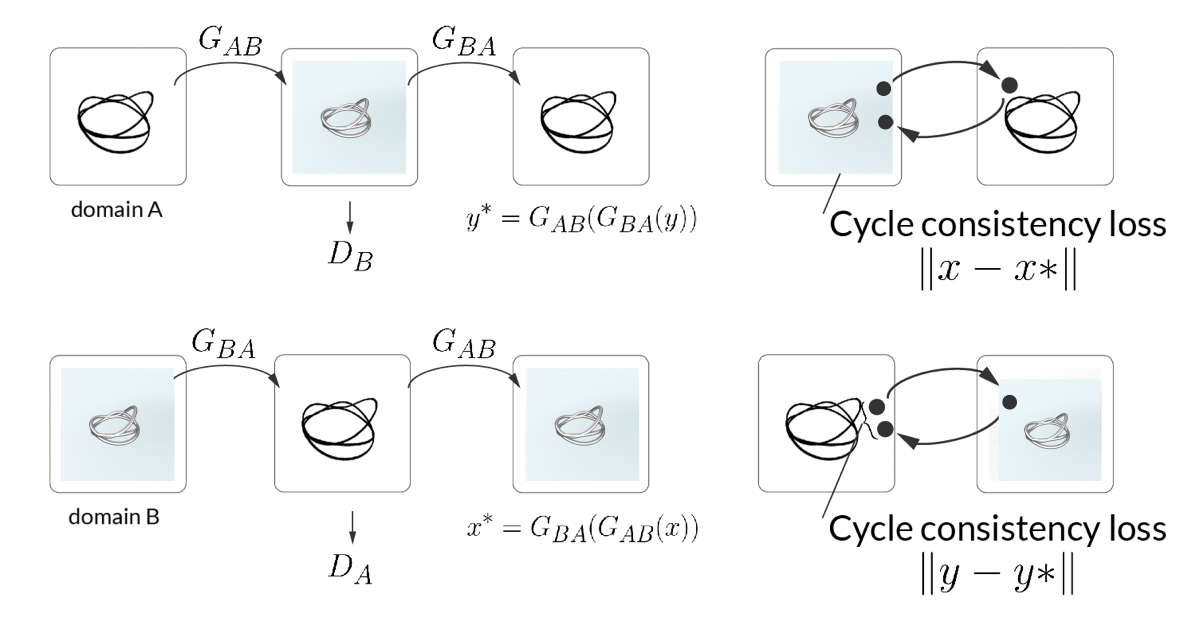} 
    \caption{Proposed CycleGAN model to learn an unsupervised Sketch2Rendering mapping.}
    \label{fig:proposed CycleGAN}
\end{figure}

The residual block used in transformation stage 2 consists of a convolutional layer, where the input is added to the output of the convolution. This is done so that the characteristics of the output image (e.g., the shapes of objects) do not differ too much from the input. Figure \ref{fig:proposed CycleGAN} shows the proposed architecture with example paired images as input.

%\begin{figure}[h!]
%    \centering
%    \includegraphics[width=0.65\textwidth]{figures/generator-architecture__.jpg} 
%    \caption{Generator architecture.}
%    \label{fig:gen architecture}
%\end{figure} 

\subsubsection{CycleGAN Discriminator architecture:}
% ----------------------------------------------------------------
%         Layer (type)               Output Shape         Param #
% ================================================================
%             Conv2d-1         [-1, 64, 400, 400]           9,472
%   FeatureMapBlock-2         [-1, 64, 400, 400]               0
%             Conv2d-3        [-1, 128, 200, 200]         131,200
%          LeakyReLU-4        [-1, 128, 200, 200]               0
%   ContractingBlock-5        [-1, 128, 200, 200]               0
%             Conv2d-6        [-1, 256, 100, 100]         524,544
%     InstanceNorm2d-7        [-1, 256, 100, 100]               0
%          LeakyReLU-8        [-1, 256, 100, 100]               0
%   ContractingBlock-9        [-1, 256, 100, 100]               0
%           Conv2d-10          [-1, 512, 50, 50]       2,097,664
%   InstanceNorm2d-11          [-1, 512, 50, 50]               0
%         LeakyReLU-12          [-1, 512, 50, 50]               0
%  ContractingBlock-13          [-1, 512, 50, 50]               0
%           Conv2d-14            [-1, 1, 50, 50]             513
% ================================================================
% Total params: 2,763,393
% Trainable params: 2,763,393
% Non-trainable params: 0
% ----------------------------------------------------------------
% Input size (MB): 1.83
% Forward/backward pass size (MB): 390.64
% Params size (MB): 10.54
% Estimated Total Size (MB): 403.02
% ----------------------------------------------------------------
The discriminator of the CycleGANs is based in the PatchGAN architecture \cite{Isola_2017}. The difference between this architecture and the usual GAN's discriminators is that the CycleGAN discriminator, instead of having a single float as an output, it outputs a matrix of values. A PatchGAN architecture will output a matrix of values, each of them between 0 (fake) and 1 (real), classifying the corresponding portions of the image. %\cnat{are you sure this isnt done in more GANs? I m not certain. DONE! It's used in any GAN that uses PatchGAN architecture in the discriminator, las pix2pix tambien lo hacen creo, pero no lo hacen las vanila GAN, las primeras que se crearon}

%\begin{figure}[h!]
%    \centering
%    \includegraphics[width=0.65\textwidth]{figures/discriminator_architecture_.jpg} 
%    \caption{Discriminator architecture and example of the classification of a portion of the image in the PatchGAN architecture, part of the CycleGAN discriminator. In this example 0.8 is the score the discriminator gave to that patch of the image (i.e., it considers that this patch looks closer to be a real image (1)).
%    \label{fig:dis arch}
%\end{figure} 

\begin{figure}[h!]
    \centering
    \includegraphics[width=0.65\textwidth]{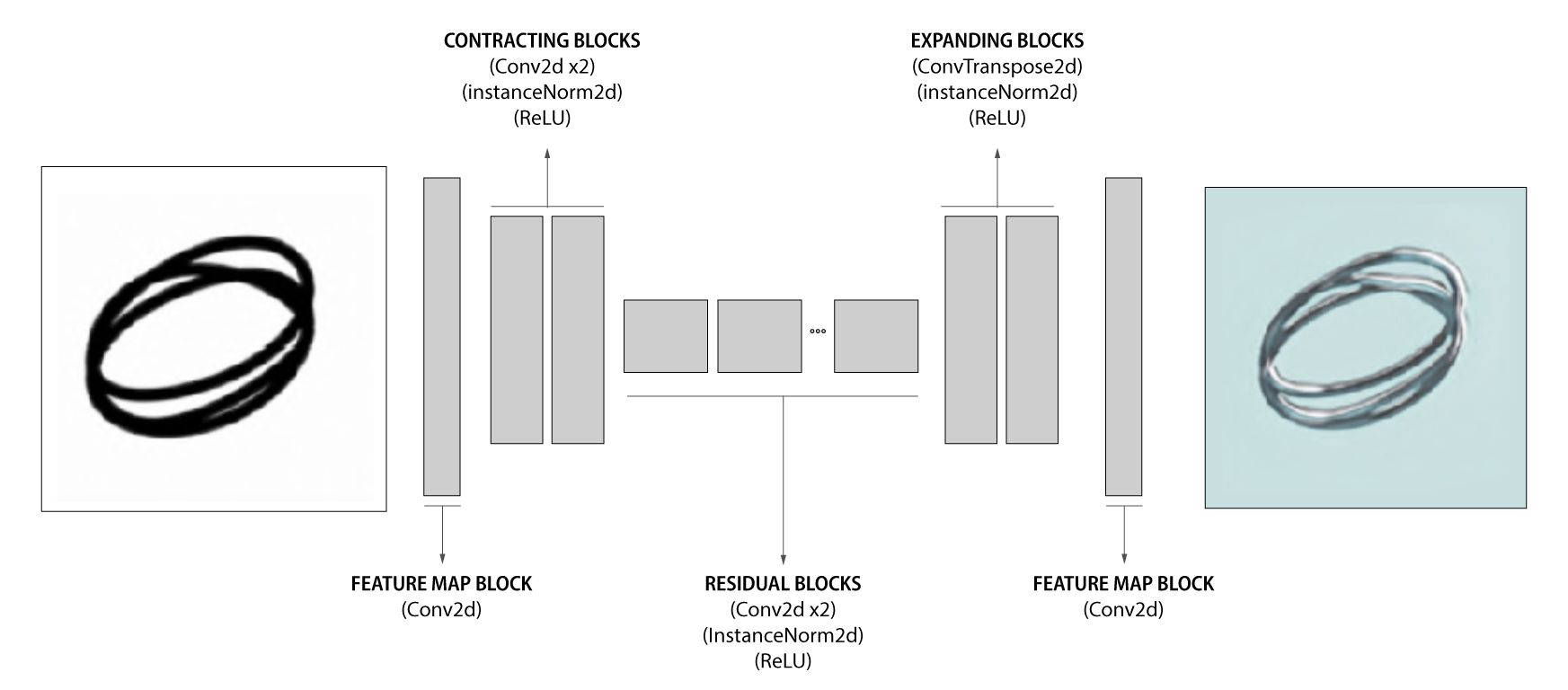} 
    \includegraphics[width=0.65\textwidth]{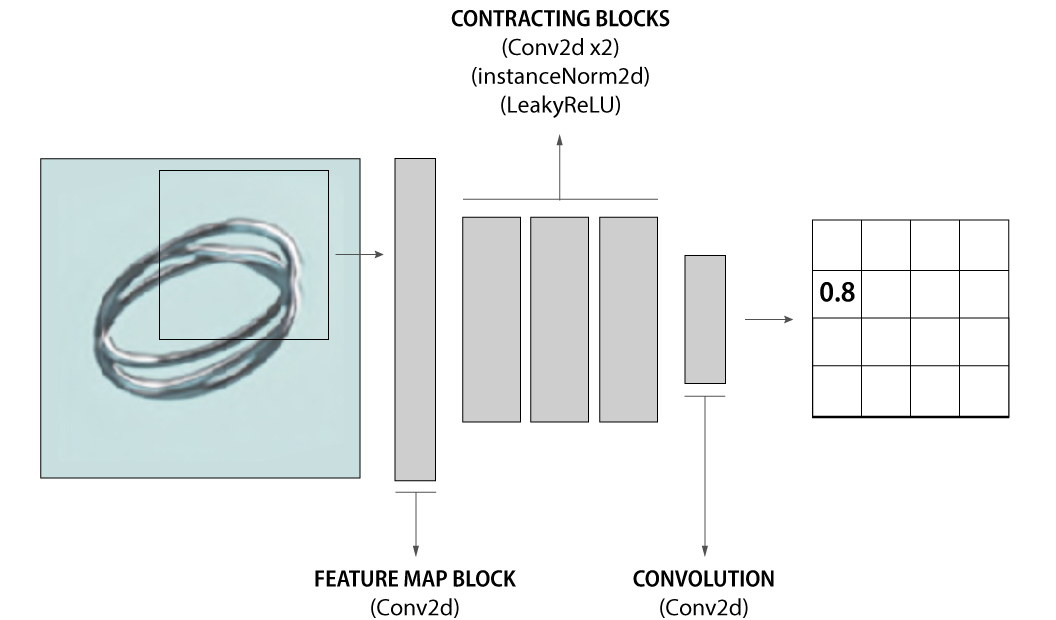} 
    \caption{Generator (upper) and Discriminator (below) architectures. Example classifying a portion of the image in the PatchGAN architecture, part of the CycleGAN discriminator. In this example 0.8 is the score the discriminator gave to that patch of the image (i.e., %it considers that 
    this patch looks closer to a real image (1)).}
    \label{fig:gen architecture}
\end{figure} 

% \subsubsection{Formulation}
% \ctomas{no estaba seguro de si ponerlo como una subsection o una subsubsection}

\subsubsection{Losses:}
The objective of CycleGANs is to learn the mapping between domains X and Y given training examples $x_i\in X$ and $y_i\in Y$. The data distributions are $x \sim p_{data}(x)$ and $y\sim p_{data}(y)$. As shown in Figure \ref{fig:proposed CycleGAN}, the model includes two mappings, one learned by each generator, $G_{AB} : X \rightarrow Y $ and $G_{BA} : Y \rightarrow X $.
%\ctomas{no estaba seguto de si llamar a los dominios X→Y o A→B, ya que los generadores y denominadores son A y B}

Apart from these generators, the model has two discriminators, one for each domain. $D_X$ will learn to distinguish between real images $x$ and fake images $x*=G_{BA}(y)$, while discriminator $D_B$ will learn to distinguish between real images $y$ and fake images $y*=G_{AB}(x)$. The objective functions will therefore contain two different losses, the adversarial losses \cite{goodfellow2014generative} that will measure whether the distribution of the generated images match the data distribution in the target domain, and the cycle consistency losses \cite{CycleGAN2017}, that will make sure that $G_{AB}$ and $G_{BA}$ do not contradict each other.

\subsubsection{Cycle consistency loss:} It can be expressed as $||x-x^*||$ or $||y-y^*||$, depending on which of the styles we consider as the starting point, where $x^*$ and $y^*$ represent the fake images generated by the generators. These equations ensure that the original image and the output image, after completing the cycle, i.e., the twice-translated image, are the same. This loss function is expressed as:
\begin{eqnarray}
\mathcal{L}_{cyc}(G_{AB},G_{BA})=&\mathbb{E}_{x\sim p_{data}(x)}[||G_{BA}(G_{AB}(x))-x||_1] \nonumber\\
                                             &+\mathbb{E}_{y\sim p_{data}(y)}[||G_{AB}(G_{BA}(y))-y||_1]
    \label{eq:cycle consistency loss}
\end{eqnarray}

%\cnat{DONE!check notation in math, you mean abs value?|}\cnat{ DONE! wher x* represents ...?  expresa cada loss function que cycleGAN use y numerala usando begin eq y end equation como para L cyc)}\cnat{Done! is or are? is this done for both domains directions or just one?}\cnat{DONE! se supone la eq sustituye a la imagen? xq no son la misma, check, y que es E de x-p? indicate}

\paragraph{Adversarial loss:} Apart from the cycle consistency loss, CycleGANs also use adversarial loss to train. As in traditional GAN models, the adversarial loss measures whether the generated images look real, i.e., whether they are indistinguishable from the ones coming from the same probability distribution learned from the training set \cite{goodfellow2014generative}. For the mapping $G_{AB} : X \rightarrow Y $ and the corresponding discriminator, we express the objective as:
\begin{eqnarray}
    \mathcal{L}_{GAN}(G_{AB},D_B,X,Y)= &&\mathbb{E}_{x\sim p_{data}(x)}[log D_B(y)] \nonumber\\
                                       &&+ \mathbb{E}_{x\sim p_{data}(x)}[log(1- D_B(G_{AB}(x))]
    \label{eq:Adversarial loss}
\end{eqnarray}

Every translation by the $G_{AB}$ generator will be checked by the $D_{B}$ discriminator, and the output of generator $G_{BA}$ will be assessed and controlled by the $D_{A}$ discriminator. Every time we translate from one domain to another, the discriminator will test if the output of the generator looks real or fake. Each generator will try to \textit{fool} its adversary, the discriminator. While each generator tries to minimize the objective function, the corresponding discriminator tries to maximize it. The training objectives of this loss are $\min_{G_{AB}}\max_{D_B}\mathcal{L}_{GAN}(G_{AB},D_B,X,Y)$ and $\min_{G_{BA}}\max_{D_A}\mathcal{L}_{GAN}(G_{BA},D_A,X,Y)$.

\paragraph{Identity loss:} The identity loss measures if the output of the CycleGAN preserves the overall color temperature or structure of the picture. Pixel distance is used to ensure that ideally there is no difference between the output and the input, this ensures that the CycleGAN only changes the parts of the image when it needs to. 

\paragraph{Model training:} The full objective of the CycleGAN is reducing these three loss functions. Actually, Zhu et al. show that training the networks with only one of the functions doesn’t arrive to high-quality results. In the formula, we can see that both the identity loss and cycle consistency functions are weighted by $\lambda_{ident}$ and $\lambda_{cyc}$, respectively. These scalars control the importance of each of the losses in the training. In our case, following the values for these parameters proposed in the original paper \cite{CycleGAN2017}, $\lambda_{cyc}$ will be 10, and $\lambda_{ident}$  will be 0.1, as this last function only controls the tint of the background of the input and output images; and as our dataset is composed of the same colors, it does not suppose a large influence.
\begin{eqnarray}
    \mathcal{L}(G_{AB},G_{BA},D_A,D_B)\hspace{-0.5mm} =\hspace{-0.5mm}&\mathcal{L}_{GAN}(G_{AB},D_B,X,Y)+\mathcal{L}_{GAN}(G_{BA},D_A,X,Y) \nonumber\\
                                        &+\lambda_{cyc}\mathcal{L}_{cyc}(G_{AB},G_{BA}) +\lambda_{ident}\mathcal{L}_ {ident}(G_{AB},G_{BA})
    \label{eq:total loss}
\end{eqnarray}

% \begin{figure}[h!]
%     \centering
%     \includegraphics[width=0.6\textwidth]{figures/image43.png} 
%     \caption{Loss functions plot for the first $1400^{th}$ steps of the training, corresponding to 14 epochs.On the image the discriminator and generator losses can be seen, as well as the individual losses that compose the total generator loss, the adversarial, cycle and identity losses.
%     }
%     \label{fig:Cycle consistency loss trainig}
% \end{figure} 
%\ctomas{he quitado esta imagen}

\subsection{CycleGAN Training details}

The networks were trained from scratch with a starting learning rate of 0.0002 for 100 epochs, after this, it was trained for 100 epochs more with a learning rate of 0.00002, as suggested by Zhu et. al in \cite{CycleGAN2017}. Following this procedure, the objective loss function of the discriminator $D$ was divided by 2, which slows down the rate at which $D$ learns compared with the generator $G$.  

For the generator and discriminator we adopt the same architectures as the ones proposed by Zhu et al. \cite{CycleGAN2017}, with the difference that for the first and last layers in the generator, we used a padding of 3 due to the input image size of our dataset.

% For the generator and discriminator we adopt the same architectures and naming conventions from Zhu et al. \cite{CycleGAN2017} and Johnson et al. \cite{johnson2016perceptual}.

% \subsubsection{Generator architectures}
% Let \texttt{c7s3-k} denote a 7x7 Convolution with \textit{k} filters and padding 3 and reflection padding. \texttt{dk} denotes a contracting block that contains a 3x3 Convolution-InstanceNorm-ReLU layer with \textit{k} filters and stride 1 and padding 2.\texttt{Rk} denotes a residual block that contains two 3×3 Convolutional layers with \textit{k} filters on each layer and stride 1 and padding 1, followed by InstanceNorm and ReLU activation. \texttt{ek} denotes a expanding block each composed of a 3x3 Transposed Convolutional layer with \textit{k} filters and stride 1 and padding 2 followed by a InstanceNorm and a ReLU activation. Tanh layer is used as a final activation layer for the generator.

% The network of the generator consists of: \texttt{c7s3-64, d128, d256, R256, R256, R256, R256, R256, R256, R256, R256, R256, u128, u64, c7s3-3}.

% \subsubsection{Discriminator architectures}

% Following Zhu et al. \cite{CycleGAN2017} we use a PatchGAN architecture \cite{Isola_2017} for our discriminator. Let \texttt{Ck} denote a 4x4 Convolution-InstanceNorm-LeakyReLU layer with \textit{k} filters and stride 2. For the LeakyReLU we use a slope of 0.2. The discriminator is composed of: \texttt{C64, C128, C256, C512}. The first layer, \texttt{C64}, does not have InstanceNorm. After the last layer, a final convolution is applied to have a 1 dimensional output.

\end{document}